\documentclass[a4paper, 10pt, conference]{ieeeconf}      
\let\labelindent\relax
\usepackage{FG2026}

\FGfinalcopy 

\IEEEoverridecommandlockouts                              
\usepackage{amsmath,graphicx,hyperref}
\usepackage{subcaption}
\usepackage{xcolor}
\usepackage{amsfonts, amssymb}

\usepackage{amsmath}
\usepackage{algorithmic}
\usepackage{diagbox}
\usepackage{arydshln}
\usepackage[linesnumbered,ruled,vlined]{algorithm2e}
\usepackage{multirow}
\usepackage{enumitem}
\usepackage{fancyhdr}

\newcommand{\updated}{\textcolor{black}} 

\def\FGPaperID{407} 

\title{\LARGE \bf
Flow Augmentation and Knowledge Distillation for Lightweight Face Presentation Attack Detection
}

\author{\parbox{16cm}{\centering
    {\large Muhammad Shahid Jabbar$^1$, Muhammad Sohail Ibrahim$^2$, Taha Hasan Masood Siddique$^3$, Kejie Huang$^3$ and Shujaat Khan$^{1,4}$}\\
    {\normalsize
    $^1$ SDAIA-KFUPM Joint Research Center for Artificial Intelligence, King Fahd University of Petroleum \& Minerals, Dhahran, Saudi Arabia\\
    $^2$ Interdisciplinary Research Center for Intelligent Secure Systems (IRC-ISS), King Fahd University of Petroleum \& Minerals, Dhahran, Saudi Arabia\\
    $^3$ College of Information Science \& Electronic Engineering, Zhejiang University, Hangzhou, China\\
    $^4$ Department of Computer Engineering, College of Computing and Mathematics, King Fahd University of Petroleum \& Minerals, Dhahran, Saudi Arabia}}
}

\begin{document}



\maketitle
\thispagestyle{fancy}

\begin{abstract}

\updated{Face presentation attack detection (FacePAD) remains challenging under diverse spoofing representation, including 2D print and replay, 3D mask-based spoofing, makeup-induced appearance manipulation, and physical occlusions, as well as under varying capture conditions.}
Motion cues are highly discriminative for FacePAD but typically require explicit optical flow estimation, which introduces substantial computational overhead and limits real-time deployment. In this work, we leverage optical flow to enhance motion representation during training while eliminating the need for flow computation at inference. We propose a dual-branch teacher model that fuses appearance cues from RGB frames with motion cues derived from colorwheel-encoded optical flow, enabling effective modeling of micro-motions and temporal consistency. To enable efficient deployment, we introduce a knowledge distillation framework that transfers motion-aware knowledge from the flow-augmented teacher to a lightweight RGB-only student via logit distillation. As a result, the student implicitly learns motion-sensitive representations without requiring explicit flow estimation or additional feature extraction blocks at inference. Extensive experiments demonstrate strong performance across multiple benchmarks, achieving 0.0\% HTER on Replay-Attack and Replay-Mobile, 0.94\% HTER on ROSE-Youtu, 5.65\% HTER on SiW-Mv2, and 0.42\% ACER on OULU-NPU. The distilled student achieves performance comparable to or better than the teacher while significantly reducing parameters and FLOPs, achieving 52 FPS on an NVIDIA\textregistered Jetson Orin Nano, indicating its suitability for real-time and resource-constrained FacePAD deployment. 

\end{abstract}


\section{INTRODUCTION}
\label{sec:intro}



Face recognition systems are widely deployed in biometric applications such as smartphone authentication, surveillance, mobile payments, and access control, yet they remain vulnerable to presentation attacks (PAs) including printed photos, replayed videos, and 2D/3D masks. Face presentation attack detection (FacePAD), also known as face anti-spoofing, aims to distinguish bonafide users from spoofing attempts \cite{9925105}.

Early FacePAD methods rely on handcrafted features and classical machine learning, however, such techniques suffered from limited robustness to variations in illumination, camera quality, and capture conditions \cite{9925105}. With the advent of deep learning, CNN-based approaches have become dominant, achieving improved generalization using learned appearance representations \cite{lucena2017transfer,nagpal2019performance,gan20173d}. To further enhance liveness detection, auxiliary cues have been explored, including physiological signals \cite{liu2018learning} and depth supervision \cite{atoum2017face, jabbar2025knowledge}. However, such approaches often incur high computational cost or remain susceptible to sophisticated spoofing artifacts.

Motion information has emerged as a particularly effective cue for FacePAD, as genuine faces exhibit non-rigid temporal dynamics that are difficult to reproduce using spoofing media. Optical flow provides a principled representation of such motion by modeling pixel-level displacement across frames. Early works exploit differences between planar and non-planar motion fields for liveness detection \cite{bao2009liveness,kollreider2009non}, while recent approaches incorporate handcrafted or learned flow-based descriptors to capture temporal dynamics \cite{yin2016face,feng2016integration}. More recent deep models integrate optical flow with appearance cues using attention mechanisms or deep dynamic texture representations to better model fine-grained facial motion \cite{shao2018joint,cheng2025integrating,li2022face}. Despite their effectiveness, such methods typically require explicit optical flow estimation and additional feature extraction blocks, leading to increased inference latency and memory overhead that limit real-time and resource-constrained deployment.

To overcome these limitations, we propose a knowledge distillation framework that leverages motion cues during training while eliminating explicit optical flow computation at inference. A motion-aware teacher model incorporates a dedicated optical flow branch to guide the learning of robust liveness representations, and transfers this knowledge to a lightweight RGB-only student model via logit-based knowledge distillation. The key contributions of this work are summarized as follows:

\begin{itemize}[leftmargin=1.2em,labelsep=0.5em,itemsep=0.2ex,topsep=0.2ex]
    \item We propose a dual-branch FacePAD teacher architecture that fuses RGB appearance cues with colorwheel-encoded optical flow to capture micro-motions and temporal consistency.
    \item We exploit consecutive video frames ($\Delta t = 1$) to enhance motion cues without introducing additional sensing modalities.
    \item We introduce a logit-based knowledge distillation strategy that enables a lightweight RGB-only student to implicitly learn motion-sensitive representations, removing the need for explicit optical flow estimation at inference.
    \item Extensive experiments demonstrate that the distilled student achieves competitive performance with significantly reduced computational complexity, enabling efficient real-time FacePAD deployment.
\end{itemize}

\section{PROPOSED METHOD}
\label{sec:methodology}

\begin{figure*}[h]
    \centering
    \includegraphics[width=0.62\linewidth]{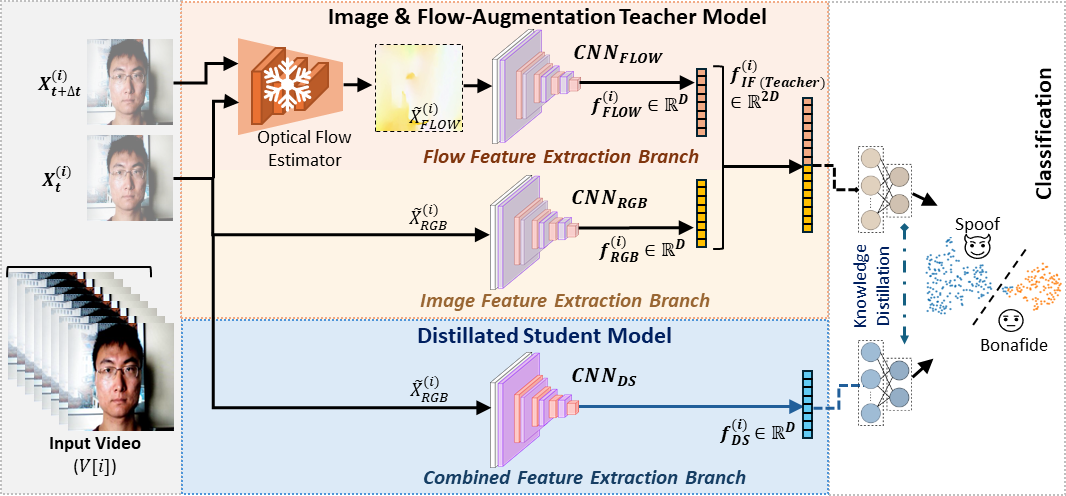}
    \caption{Proposed dual-branch (RGB+Flow) teacher and distilled RGB-only student for FacePAD.}
    \label{fig:DualBranchArchitecture}
\end{figure*}

\subsection{Frame Sampling and Flow Estimation}
\label{sec:preproc}

The input consists of RGB video frames. From each clip, a reference frame at time \(t\),

\[X_{t}^{(i)} \in \mathbb{R}^{3 \times H \times W},\]

is sampled, where \(i\) indexes the clip. During training, \(t\) is chosen randomly to promote variability; at inference, the first frame is used. To extract motion cues, a temporally adjacent frame at \(t+\Delta t\) (where $\Delta t=1$),
\[
X_{t+\Delta t}^{(i)} \in \mathbb{R}^{3 \times H \times W},
\]
is paired with \(X_{t}^{(i)}\), and optical flow is estimated on-the-fly:
\[
\mathbf{U}_{t,t+\Delta t}^{(i)} = (u,v) \;\gets\; \text{FlowEngine}\!\left(X_{t}^{(i)},\, X_{t+\Delta t}^{(i)}\right) \in \mathbb{R}^{2 \times H \times W}.
\]

\textbf{Flow estimation.} The FlowEngine is instantiated with the \textit{UniMatch} network \cite{unimatch}, using the MixData-pretrained checkpoint and standard hyperparameters (multi-scale transformer encoder, iterative refinement, and windowed attention). To match the training distribution of UniMatch, inputs are scaled to the 0–255 range if originally in \([0,1]\), transposed if necessary to ensure width \(\geq\) height, and bilinearly resized to multiples of 32. The resulting flow field \(\mathbf{U}_{t,t+\Delta t}^{(i)}\) is resized back to the original resolution with magnitude correction, and any transposition is reversed. The estimator outputs pixel-wise displacements in \emph{pixel units}, aligned with the original grid.

\textbf{Pre-processing and augmentation.} The 2-channel flow field \(\mathbf{U}_{t,t+\Delta t}^{(i)}\) is encoded into a 3-channel image via HSV-to-RGB colorwheel mapping,
\[
X_{\text{FLOW},t}^{(i)} \in \mathbb{R}^{3 \times H \times W},
\]
for compatibility with CNN backbones. Identical geometric operations are then applied to both modalities:  
(i) adaptive center cropping to the largest square region, and  
(ii) resizing to \(\rho \times \rho\) with \(\rho=224\).  
When resizing flow, vector magnitudes are scaled to preserve physical consistency. During training, synchronized random augmentations (in-plane rotation, isotropic scaling) are applied with shared parameters to maintain alignment between \(X_{t}^{(i)}\) and \(\mathbf{U}_{t,t+\Delta t}^{(i)}\). At inference, only deterministic cropping and resizing are used.

For knowledge distillation, the student model follows the same preprocessing pipeline but requires only RGB frames, eliminating dependence on the flow estimator at inference.

\subsection{Network Architecture}
\label{sec:method_arch}


The proposed approach couples appearance and motion cues through a dual-branch teacher network fed by an RGB frame and a colorwheel-encoded flow image computed from an adjacent frame (Algorithm \ref{alg:inference_imgflow}; Fig. \ref{fig:DualBranchArchitecture}). After preprocessing, each sample provides
\(\tilde{X}_{\text{RGB}}^{(i)} \in \mathbb{R}^{3 \times \rho \times \rho}\) and
\(\tilde{X}_{\text{FLOW}}^{(i)} \in \mathbb{R}^{3 \times \rho \times \rho}\).

\SetKwInput{KwGiven}{Given}
\begin{algorithm}[h!]
\caption{FacePAD teacher model with flow augmentation}
\label{alg:inference_imgflow}
\small
\SetAlgoLined
\KwIn{Batch of video sequences \( V \in \mathbb{R}^{B \times 3 \times H \times W} \)}
\KwOut{Class predictions \( Y \in \mathbb{R}^{B \times 2} \)}

\KwGiven{\\
\quad \(\textbf{FlowEngine}\): optical flow estimator\;
\quad \(\textbf{Colorwheel}\): mapping flow \((u,v)\) to RGB via HSV\;
\quad \(\textbf{SyncAug}\): synchronized transforms for RGB and Flow\;
\quad \(\text{\textbf{CNN}}_{\text{\textbf{RGB}}},\, \text{\textbf{CNN}}_{\text{\textbf{FLOW}}}\): encoders\;
\quad \(\textbf{FC}\): classifier head\;
\quad \textbf{Temporal offset}: \(\Delta t =1\)}

Initialize an empty list \( Y = [\,] \).

\For{each sample \( i = 1,\dots,B \)}{
    Reference frame: \( X_{t}^{(i)} \leftarrow V[i] \). \\
    Adjacent frame: \( X_{t+\Delta t}^{(i)} \). \\
    Flow: \( \mathbf{U}_{t,t+\Delta t}^{(i)} \leftarrow \textbf{FlowEngine}(X_{t}^{(i)}, X_{t+\Delta t}^{(i)}) \). \\
    Encode as RGB: \( X_{\text{FLOW},t}^{(i)} \leftarrow \textbf{Colorwheel}(\mathbf{U}_{t,t+\Delta t}^{(i)}) \). \\
    Preprocess: \((\tilde{X}_{\text{RGB}}^{(i)},\, \tilde{X}_{\text{FLOW}}^{(i)}) \leftarrow \textbf{SyncAug}(X_{t}^{(i)}, X_{\text{FLOW},t}^{(i)}) \). \\
    Features: 
    \\ \quad \( \mathbf{f}_{\text{RGB}}^{(i)} \leftarrow \text{CNN}_{\text{RGB}}(\tilde{X}_{\text{RGB}}^{(i)}) \), 
    \\ \quad \( \mathbf{f}_{\text{FLOW}}^{(i)} \leftarrow \text{CNN}_{\text{FLOW}}(\tilde{X}_{\text{FLOW}}^{(i)}) \). \\
    Concatenate: \( \mathbf{f}^{(i)} \leftarrow [\mathbf{f}_{\text{RGB}}^{(i)},\, \mathbf{f}_{\text{FLOW}}^{(i)}] \). \\
    Predict: \( y^{(i)} \leftarrow \text{FC}(\mathbf{f}^{(i)}) \); append to \(Y\).
}
\Return{Predictions \( Y \)}
\end{algorithm}
\normalsize

Two ImageNet-pretrained MobileNetV3-Large backbones encode RGB and flow inputs independently. {MobileNetV3-Large backbones are used for its favorable performance–efficiency trade-off, which is critical for real-time FacePAD deployment on edge and embedded platforms \cite{jabbar2025knowledge, khan2025spatio, siddique2025advspoofguard, ibrahim2025improving}}.
\[
\mathbf{f}_{\text{RGB}}^{(i)} = \mathrm{CNN}_{\text{RGB}}(\tilde{X}_{\text{RGB}}^{(i)}), \quad
\mathbf{f}_{\text{FLOW}}^{(i)} = \mathrm{CNN}_{\text{FLOW}}(\tilde{X}_{\text{FLOW}}^{(i)}).
\]
The flow branch processes the 3-channel colorwheel-encoded image, implicitly capturing motion direction and magnitude. Late fusion concatenates features,
\[
\mathbf{f}^{(i)} = \big[\,\mathbf{f}_{\text{RGB}}^{(i)},\, \mathbf{f}_{\text{FLOW}}^{(i)}\,\big],
\]
followed by an MLP classifier producing logits \(\mathbf{s}^{(i)} \in \mathbb{R}^{2}\) and posterior scores \(\hat{\mathbf{p}}^{(i)}=\mathrm{softmax}(\mathbf{s}^{(i)})\).

\textbf{Knowledge distillation.} We distill the dual-branch teacher into a single-branch MobileNetV3-Large student operating on RGB only. The student is trained with a combined objective:

\[
\begin{aligned}
\mathcal{L}
&= (1-\alpha)\,\mathrm{CE}(y, s_S) \\
&\quad + \alpha\,T^2\,\mathrm{KL}\!\left(
\mathrm{softmax}\!\left(\tfrac{s_T}{T}\right)
\,\big\|\,
\mathrm{softmax}\!\left(\tfrac{s_S}{T}\right)
\right),
\end{aligned}
\]


where \(\mathrm{CE}\) is the cross-entropy between student logits \(s_S\) and ground-truth label \(y\), \(\mathrm{KL}\) matches softened distributions of teacher and student logits, \(T\) is the temperature, and \(\alpha\) balances the terms. This eliminates reliance on flow estimation at inference while retaining competitive performance.

\subsection{Performance Metrics}
\label{metrics}

We report both \emph{classification} and \emph{computational} metrics. Classification metrics include Accuracy, AUC-ROC, and Half Total Error Rate (HTER). HTER is defined as 
\(\text{HTER} = (\text{FAR}+\text{FRR})/2\), 
with \(\text{FAR} = FP/(FP+TN)\) and \(\text{FRR} = FN/(FN+TP)\),
where \(TP\), \(TN\), \(FP\), and \(FN\) denote true/false positives and negatives.

AUC-ROC is computed by threshold sweeping. We report Equal Error Rate (EER), defined at the threshold where \(\text{FAR}=\text{FRR}\), and the Youden index \(\max(\text{TPR}-\text{FPR})\) as a complementary summary. We also report attack classification error rate (ACER), defined as \(\text{ACER} = \frac{\text{APCER + BPCER}}{2}\), where APCER and BPCER are defined as attack presentation classification error rate and bonafide presentation classification error rate, respectively.

Computational efficiency is reported with model size (parameters) and FLOPs per \(\rho\times\rho\) input to compare the dual-branch teacher and distilled student, along with deployment throughput (fps) for the student model.


\subsection{Implementation Details}
The flow-augmented teacher is trained using Adam (lr=$10^{-4}$) with cross-entropy loss, batch size 16, for up to 100 epochs, with an early stopping patience of 20 epochs. 
Testing is performed with batch size 256. 
Optical flow is computed \emph{on-the-fly} using UniMatch (Section \ref{sec:preproc}) with the temporal offset of $\Delta t =1$. 
The distilled student (DS) employs MobileNetV3-Large with RGB-only input, trained under the same optimization settings, with distillation temperature $T{=}3$ and balance factor $\alpha{=}0.7$. 
All experiments are conducted on a single NVIDIA RTX A4500 GPU. 
Three independent experimental runs are conducted, and mean results are reported.


\section{EXPERIMENTS AND RESULTS}
\label{sec:expresults}

\subsection{Datasets}
We evaluate the proposed models on three widely used FacePAD benchmarks: \textsc{Replay-Attack} (RA) \cite{chingovska2012effectiveness}, \textsc{Replay-Mobile} (RM) \cite{costa2016replay}, \textsc{ROSE-Youtu} (RY) \cite{Li2018UnsupervisedDA}, \textsc{OULU-NPU} \cite{boulkenafet2017oulu}, and SiW-Mv2 \cite{guo2022multi}. 
RA, RM, and OULU-NPU comprise bonafide and spoof attempts with print and video-replay attacks, RY additionally includes paper-mask attacks and multiple capture devices, while SiW-Mv2 includes a 14 different attacks, comprising obfuscation makeup, partial coverings, etc., making it particularly challenging. 
We follow the official subject-disjoint protocols. 
From each video clip, a single RGB reference frame is sampled along with its temporal neighbor at offset $\Delta t=1$ for flow-based experiments. 
Table \ref{tab:dataset} summarizes dataset statistics.

\begin{table}[h]
    \centering
    \caption{Summary of datasets used in this study.}
    \resizebox{0.9\linewidth}{!}{
    \begin{tabular}{l | c c c}
        \hline
        \textbf{Dataset} & \textbf{Year} & \textbf{Subjects} & \textbf{Genuine / Spoof Clips} \\
        \hline
        Replay-Attack \cite{chingovska2012effectiveness} & 2012 & 50 & 300 / 1000 \\
        Replay-Mobile \cite{costa2016replay} & 2016 & 39 & 540 / 624 \\
        ROSE-Youtu \cite{Li2018UnsupervisedDA} & 2018 & 20 & 1000 / 2350 \\
        OULU-NPU \cite{boulkenafet2017oulu} & 2017 & 55 & 990 / 3960 \\
        SiW-Mv2 \cite{guo2022multi} & 2022 & 600 & 785 / 915 \\
        \hline
    \end{tabular}}
    \label{tab:dataset}
\end{table}



\subsection{Results}

\begin{table*}[t]
\centering
\caption{Performance comparison of the proposed flow-augmented FacePAD method ($I\&F$) with the image-only model ($I$) on the benchmark datasets.}
\label{tab:flow_results}
\resizebox{0.8\linewidth}{!}{%
\begin{tabular}{l|cc|cc|cc|cc|cc}
\hline
\multicolumn{1}{c|}{\multirow{2}{*}{\diagbox[width=14em]{\textbf{Metric}}{\textbf{Dataset / Method}}}} &
\multicolumn{2}{c|}{\textsc{Replay-Attack}} &
\multicolumn{2}{c|}{\textsc{Replay-Mobile}} &
\multicolumn{2}{c|}{\textsc{ROSE-Youtu}} &
\multicolumn{2}{c|}{\textsc{OULU-NPU}} &
\multicolumn{2}{c}{\textsc{SiW-Mv2}} \\
\cline{2-11}
\multicolumn{1}{c|}{} &
\textbf{$I$} & \textbf{$I$ \& $F$} &
\textbf{$I$} & \textbf{$I$ \& $F$} &
\textbf{$I$} & \textbf{$I$ \& $F$} &
\textbf{$I$} & \textbf{$I$ \& $F$} &
\textbf{$I$} & \textbf{$I$ \& $F$} \\
\hline

Accuracy (\%)     & 100 & 100 & 100 & 100 & 98.06 & 98.57 & 98.89 & 99.78 & 92.30 & 91.81 \\
AUC-ROC           & 1.000 & 1.000 & 1.000 & 1.000 & 0.999 & 0.9993 & 1.000 & 0.999 & 0.981 & 0.987 \\
EER (\%)          & 0.0 & 0.0 & 0.0 & 0.0 & 1.39 & 0.98 & 0.97 & 0.14 & 6.63 & 6.08 \\
HTER (\%)         & 0.0 & 0.0 & 0.0 & 0.0 & 1.47 & 0.94 & 0.63 & 0.21 & 6.11 & 5.65 \\
FAR (\%)          & 0.0 & 0.0 & 0.0 & 0.0 & 1.39 & 0.92 & 0.42 & 0.14 & 8.01 & 7.46 \\
FRR (\%)          & 0.0 & 0.0 & 0.0 & 0.0 & 1.56 & 0.97 & 0.83 & 0.28 & 4.21 & 3.83 \\
Youden's Index    & 1.00 & 1.00 & 1.00 & 1.00 & 0.971 & 0.981 & 0.988 & 0.996 & 0.877 & 0.887 \\
ACER (\%)         & - & - & - & - & - & - & 1.25 & 0.42 & - & - \\
\hline
\end{tabular}%
}
\end{table*}

Table \ref{tab:flow_results} reports the performance of image-only ($I$) and flow-augmented ($I\&F$) models. 
On RA and RM, the flow-augmented model as well as the image-only model achieve \textbf{perfect detection} (0.0\% HTER, 100\% Accuracy). 
On RY, augmenting RGB with flow yields significant performance gain: $I \& F$ model attains improved HTER (0.94\%), AUC (0.9993), and Accuracy (98.57\%). On OULU-NPU, the \(I\&F\) model achieves significantly superior ACER (0.42\%). Similar performance trend is observed with SiW-Mv2 dataset attaining a {competitive EER (6.07\%)} compared to the performance of image-only model. 

\begin{table}[h]
    \centering
    \caption{Knowledge distillation on \textsc{ROSE-Youtu} dataset: $I\&F$ teacher vs. distilled student ($DS$).}
    \resizebox{0.7\linewidth}{!}{
    \begin{tabular}{l | c c}
        \hline
        \textbf{\diagbox[width=12em]{Metric}{Model}} & \textbf{$I\&F$}& \textbf{DS (RGB-only)} \\
        \hline
        Parameters (M) & 14.29 & \textbf{3.46} \\
        GFLOPs & 356.63 & \textbf{0.22} \\
        Model Size (MB) & 54.97 & \textbf{13.42} \\
        Peak GPU Mem. (MB) & 751 & \textbf{30} \\ \hdashline
        Accuracy (\%) & 98.57 & \textbf{98.86} \\
        AUC-ROC & 0.9993 & \textbf{0.9996} \\
        EER (\%) & \textbf{0.98} & 1.08 \\
        HTER (\%) & 0.94 & \textbf{0.81} \\
        \hline
    \end{tabular}}
    \label{tab:transfer_learning}
\end{table}

Knowledge distillation further compresses the model. As shown in Table \ref{tab:transfer_learning}, the distilled student achieves superior performance (HTER 0.81\%, AUC 0.9996, Acc. 98.86\%) compared to the teacher \(I\&F\) model while reducing parameters, FLOPs, and memory footprint by an order of magnitude. {Furthermore, implementation of the distilled student model on NVIDIA\textregistered Jetson Orin Nano resulted in a frame rate, measured in frames per second (FPS), of 52 compared to an FPS of 46 on an Intel\textregistered Core\texttrademark i5 10400@2.80 GHz CPU}. The computational performance underscores the practicality and real-time applicability of the proposed motion-cue distillation for real-world deployment.



\begin{figure}[h]
    \centering
    \renewcommand{\arraystretch}{0.5} 
    \setlength{\tabcolsep}{1pt}       
    \begin{tabular}{@{}c c c c@{}}
    
    \includegraphics[width=0.20\linewidth]{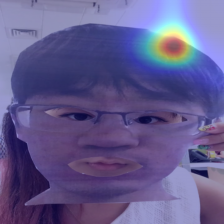} &
    \includegraphics[width=0.20\linewidth]{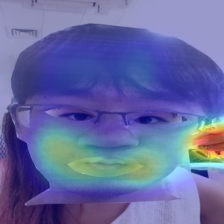} &
    \includegraphics[width=0.20\linewidth]{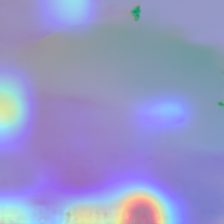} &
    \includegraphics[width=0.20\linewidth]{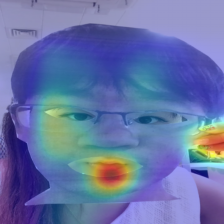} \\
    
    \includegraphics[width=0.20\linewidth]{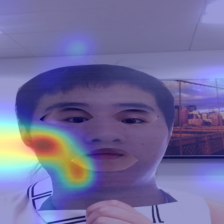} &
    \includegraphics[width=0.20\linewidth]{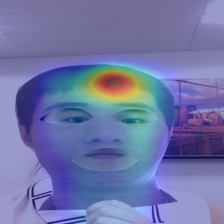} &
    \includegraphics[width=0.20\linewidth]{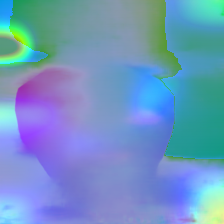} &
    \includegraphics[width=0.20\linewidth]{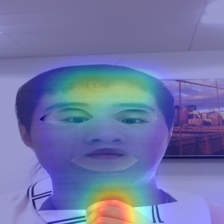} \\
    
    $I$ &
    $I\&F$ &
    $I\&F$ &
    $DS$ \\
     & (Image) & (Flow) & \\
    
    \end{tabular}
    
    \caption{Grad-CAM overlays for Image-only ($I$), flow-augmented ($I\&F$), and distilled student ($DS$) models, showing sharper motion-aware activations with flow guidance.}
    \label{fig:gradCAM}
\end{figure}
Grad-CAM visualizations (Fig. \ref{fig:gradCAM}) demonstrate that flow-guided training sharpens focus on motion-consistent regions, supporting the observed performance gains.

\subsection{Comparison with State-of-the-Art}

Table \ref{tab:comp_all} benchmark our models against recent state-of-the-art methods. On RA and RM, both teacher and student models achieve perfect detection (0.0\% HTER), matching or exceeding the best reported results. 
On RY, our distilled student achieves 0.81\% HTER, outperforming strong baselines such as CA-FAS \cite{long2024confidence}, AdvSpoofGuard \cite{siddique2025advspoofguard}, and  Depth-augmented Teacher \cite{jabbar2025knowledge} while remaining substantially more efficient.
\updated{On SiW-Mv2, the proposed teacher achieves competitive HTER/EER, while the distilled student closely matches teacher performance. Although the proposed DS model shows slightly inferior performance compared to ResNet50V2 \cite{alassafi2023fully} and EfficientNet-B0 \cite{huszar2024securing}, these methods rely on substantially larger backbones (25.6M and 5.3M parameters, respectively) compared to only 3.46M parameters for the proposed DS model.}

\begin{table}[!h]
    \centering
    \caption{Comparison with state-of-the-art methods.}

    \resizebox{\linewidth}{!}{
    \begin{tabular}{l c c c c c}
        \hline
        \textbf{Method} & \textbf{Year} & \textbf{RA} & \textbf{RM} & \textbf{RY} & \textbf{SiW-Mv2} \\
        \hline
        ResNet50V2 \cite{alassafi2023fully} & 2023 & 0.03 & 0.0 & 2.53/2.64 & \textbf{5.18}/\textbf{6.08} \\
        EfficientNet-B0 \cite{huszar2024securing} & 2024 & 36.88 & 4.62 & 9.54/- & 4.82/5.80 \\
        Spatio-Temporal ($\tau = 1$) \cite{khan2025spatio} & 2025 & 0.0 & 0.0 & 1.47/\textbf{0.85} & 6.11/6.63 \\
        AdvSpoofGuard \cite{siddique2025advspoofguard} & 2025 & 0.0 & 0.0 & 1.97/1.08 & - \\
        Depth-aug. Teacher \cite{jabbar2025knowledge} & 2025 & 0.0 & 0.0 & 1.02/1.15 & 5.82/6.08 \\
        Depth-aug. Student \cite{jabbar2025knowledge} & 2025 & 0.0 & 0.0 & 1.01/1.39 & - \\
        \textbf{Proposed (I\&F)} & 2026 & \textbf{0.0} & \textbf{0.0} & 0.94/0.98 & 5.65/6.07 \\
        \textbf{Proposed (DS)} & 2026 & \textbf{0.0} & \textbf{0.0} & \textbf{0.81}/1.08 & 5.78/6.35 \\
        \hline
    \end{tabular}}
    \vspace{1mm}
\parbox[t]{\linewidth}{
  \scriptsize{\raggedright
  \textit{Results are HTER (\%) for RA \& RM, HTER/EER (\%) for RY \& SiW-Mv2.}}}
    \label{tab:comp_all}
\end{table}

Table \ref{tab:OULUmetricfull} presents the comparative performance on OULU-NPU dataset. The proposed methods achieve superior performance compared to the state-of-the-art methods, with the distilled student model ($DS$) obtaining the best performance at 0.35\% ACER, followed by the proposed flow-augmented model ($I\&F$) with 0.42\% ACER. These results highlight the effectiveness of the proposed method for robust and lightweight FacePAD.

\begin{table}[h]
\caption{Performance comparison of the proposed method with the state-of-the-art methods on OULU-NPU dataset.}
\label{tab:OULUmetricfull}
\resizebox{1\columnwidth}{!}{%
\begin{tabular}{l|c|c|c|c}
\hline
\bf{Method} & \bf{Year} & \bf{APCER (\%)} & \bf{BPCER (\%)} & \bf{ACER (\%)} \\ \hline
ED-LBP (VAR) \cite{shu2021face} & 2021 & 11.3 & 8.4 & 9.9 \\ 
Texture (VAR) \cite{daniel2021texture} & 2021 & 14.5 & 15 & 14.8 \\ 
Fake-Net (VAR) \cite{alshaikhli2021face} & 2021 & 5.4 & 6.9 & 6.2 \\ 
OFT (VAR) \cite{li2022face} & 2022 & 5.7 & 2.7 & 4.2 \\ 
ELA \cite{lee2023face} & 2023 & - & - & 4.86 \\
3D-LCN \cite{ning2024face} & 2024 & 1.5 & \textbf{0.5} & 1.0 \\
Spatio-Temporal \cite{khan2025spatio} & 2025 & 1.94 & 0.55 & 1.25 \\ 
Depth-aug. Student (MB-V3) \cite{jabbar2025knowledge} & 2025 & 1.39 & 0.83 & 1.11 \\
Depth-aug. Student (MB-V2) \cite{jabbar2025knowledge} & 2025 & 0.28 & 0.83 & 0.56 \\

\textbf{Proposed (I\&F)} & 2026 & 0.28 & 0.56 & 0.42 \\
\textbf{Proposed (DS)} & 2026 & \textbf{0.14} & 0.56 & \textbf{0.35} \\

\hline
\end{tabular}
} 
\parbox[t]{\linewidth}{
  \scriptsize{\raggedright
  \textit{VAR: Video Attack Return, OFT: Optical Flow + Texture, MB: MobileNet.}}}

\end{table}


\section{CONCLUSION}
\label{sec:conclusion}

This paper presents an efficient FacePAD framework that leverages motion information during training to enhance liveness modeling while avoiding the computational overhead of explicit motion estimation at inference. A dual-branch teacher network integrates appearance cues from RGB frames with motion cues derived from colorwheel-encoded optical flow, enabling effective learning of micro-motion patterns and temporal consistency. Extensive experiments demonstrate strong performance across multiple benchmarks, achieving 0.0\% HTER on Replay-Attack and Replay-Mobile, 0.94\% HTER on ROSE-Youtu, 5.65\% HTER on SiW-Mv2, and 0.42\% ACER on OULU-NPU. To support deployment in resource-constrained and real-time scenarios, the motion-aware teacher is distilled into a lightweight RGB-only student model. The distilled student preserves motion-sensitive representations without requiring explicit flow computation and additional feature extraction at inference, attaining performance comparable to or better than the teacher while significantly reducing model complexity. Moreover, the student achieves 52 FPS on an NVIDIA\textregistered Jetson Orin Nano, demonstrating the practicality of the proposed motion-guided knowledge distillation framework for accurate, real-time, and deployable FacePAD systems.




\section{ACKNOWLEDGMENTS}

Taha Hasan Masood Siddique and Kejie Huang would like to acknowledge the support received from Zhejiang Province's Leading Talent Project in Science and Technology Innovation (2023R5204).

Shujaat Khan acknowledges the support of King Fahd University of Petroleum and Minerals (KFUPM) under the Early Career Grant No. EC241027.




{\small
\bibliographystyle{IEEEtran}
\bibliography{egbib}
}

\end{document}